\documentclass[10pt,twocolumn,letterpaper]{article}

\usepackage{cvpr}
\usepackage{times}
\usepackage{epsfig}
\usepackage{graphicx}
\usepackage{amsmath}
\usepackage{amssymb}

\usepackage{graphicx}
\usepackage{caption,subcaption}

\usepackage[pagebackref=true,breaklinks=true,letterpaper=true,colorlinks,bookmarks=false]{hyperref}

\cvprfinalcopy 

\def\cvprPaperID{****} 

\begin{document}

\title{Ridiculously Fast Shot Boundary Detection with Fully Convolutional Neural Networks}

\author{Michael Gygli\\
gifs.com \\
Zurich, Switzerland\\
{\tt\small michael@gifs.com}
}

\maketitle

\begin{abstract}
Shot boundary detection (SBD) is an important component of many video analysis tasks, such as action recognition, video indexing, summarization and editing.
Previous work typically used a combination of low-level features like color histograms, in conjunction with simple models such as SVMs.
Instead, we propose to learn shot detection end-to-end, from pixels to final shot boundaries.
For training such a model, we rely on our insight that all shot boundaries are generated.
Thus, we create a dataset with one million frames and automatically generated transitions such as cuts, dissolves and fades.
In order to efficiently analyze hours of videos, we propose a Convolutional Neural Network (CNN) which is fully convolutional in time, thus
allowing to use a large temporal context without the need to repeatedly processing frames.
With this architecture our method obtains state-of-the-art results while running at an unprecedented speed of more than 120x real-time.
\end{abstract}

\section{Introduction}
\begin{figure}
  \begin{subfigure}[t]{1\linewidth}
 \includegraphics[width=0.49\linewidth]{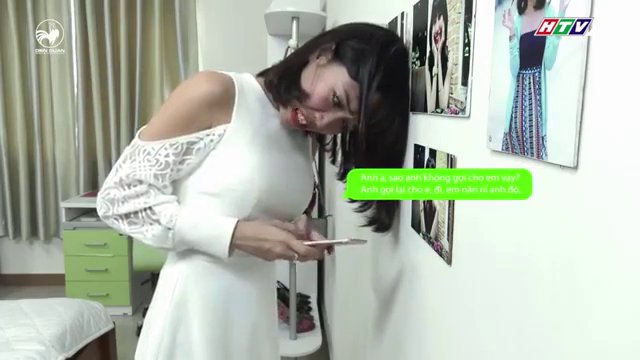}
 \includegraphics[width=0.49\linewidth]{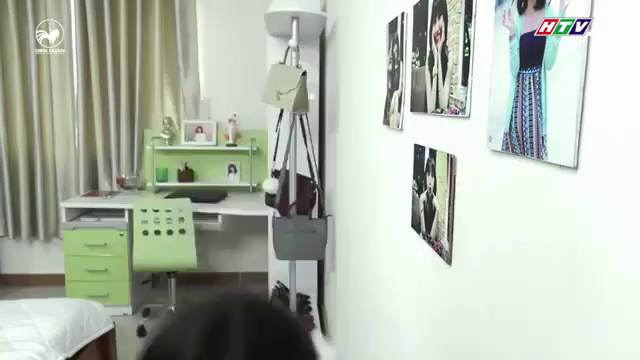} 
    \caption{Two adjacent frames from different shots.}
    \end{subfigure}
  \begin{subfigure}[t]{1\linewidth}
 \includegraphics[width=0.49\linewidth]{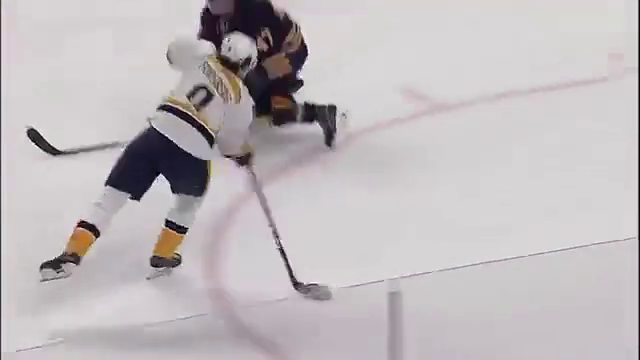}
 \includegraphics[width=0.49\linewidth]{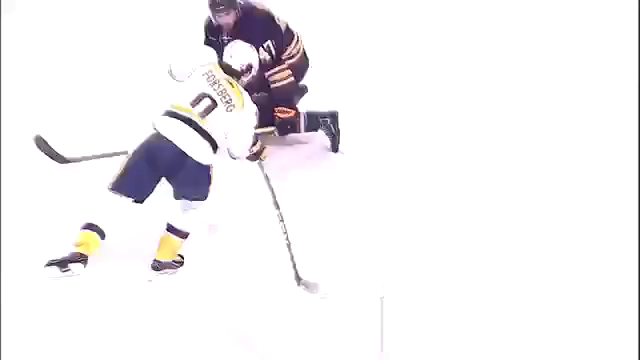}  
    \caption{Adjacent frames with strong variation due to a flash, but coming from the same shot.}
    \label{fig:hard_cases_flash}
  \end{subfigure}    
\begin{subfigure}[t]{1\linewidth}
 \includegraphics[width=0.49\linewidth]{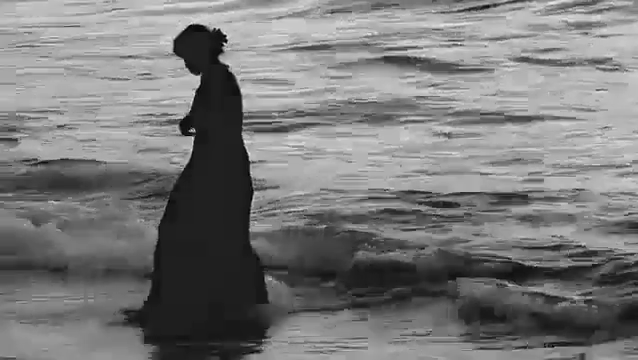}
 \includegraphics[width=0.49\linewidth]{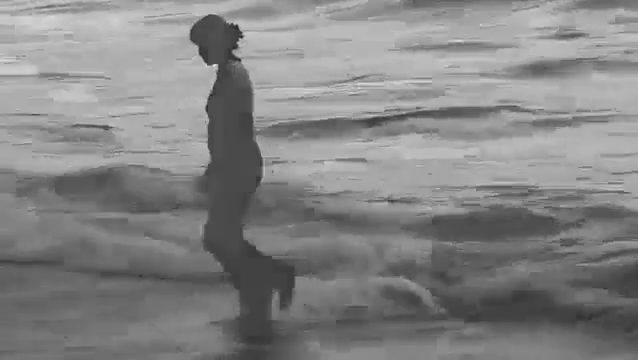} 
    \caption{Frames of a dissolve transition which are $0.5$ seconds apart, but visually very similar.}
  \end{subfigure}%
  \caption{The challenges of shot detection. Understanding if a scene shows strong variation or if a shot change occurs is often difficult.}
  \label{fig:hard_cases}
\end{figure}

A shot of a video consists of consecutive frames which show a continuous progression of video and which are thus interrelated.
The goal of shot boundary detection is to predict when such a shot starts or ends. Thus, it needs to detect transitions such as cuts, dissolves and fades.
Representing a video as a set of shots is useful for many tasks and has been an important pre-processing step for automatic video analysis such as action recognition~\cite{wang2017untrimmednets} and video summarization~\cite{gygli2015video}.
It is also useful when manually re-editing videos.

Due to the broad use of SBD, it has been researched for more than 25 years~\cite{akutsu1992video} and many methods have been proposed,~\eg~\cite{akutsu1992video,cernekova2006information,apostolidis2014fast, baraldi2015shot}.
Typical methods approach SBD with a set of low-level features, such as color and edge histogram difference or SURF~\cite{bay2008speeded},
in combination with Support Vector Machines (SVMs)~\cite{smeaton2010video,apostolidis2014fast}.
To improve the comparison of different methods and boost process, the TRECVid initiative hosted a shot detection challenge for several years~\cite{smeaton2010video}.
Nonetheless, shot detection is not solved yet. While it may appear to be a simple problem, as humans can easily spot most shot changes, it is challenging for an algorithm.
This is due to several reasons.
First, video editors often try to make shot transitions subtle, so that they are not distracting from the video content.
In some cases the transitions are completely hidden, such as in the movie \textit{Rope} by Alfred Hitchcock~\cite{hitchcock2001rope}.
Second, videos show strong variations in content and motion speed. In particular fast motion, leading to motion blur, is often falsely considered a shot change.
Third, transitions such as dissolves have small frame to frame changes, making it difficult them with traditional methods.
Figure~\ref{fig:hard_cases} shows some challenging cases.

To tackle these challenges we propose to learn shot boundary detection end-to-end, by means of a fully convolutional neural network.
In order to train this network we create a new dataset with one million frames and automatically generated labels: transition or none-transition frames.
For our dataset we generated hard cuts, crop cuts, dissolves, wipes and fade transitions. Additionally we generated none-transition examples,
that come from the same shot, including frames where we added an artificial flash. This allows to make the network invariant to flashes,
which were previously often falsely detected as cuts and corrected in a post-processing step~\cite{smeaton2010video}.

In summary, we make the following contributions:
\begin{enumerate}
\item A way to generate a large-scale dataset for training shot detection algorithm without the need to manually annotate them.
\item A novel and highly efficient CNN architecture by making it fully-convolutional in time, inspired by fully convolutional architectures for image segmentation~\cite{long2015fully}.
Our method runs at 121x real-time speed on a Nvidia K80 GPU.
\item An empirical comparison to previous state of the art in terms of accuracy and speed. We show that we improve upon existing methods in both regards.
\end{enumerate}

\begin{figure}
 \includegraphics[width=1\linewidth]{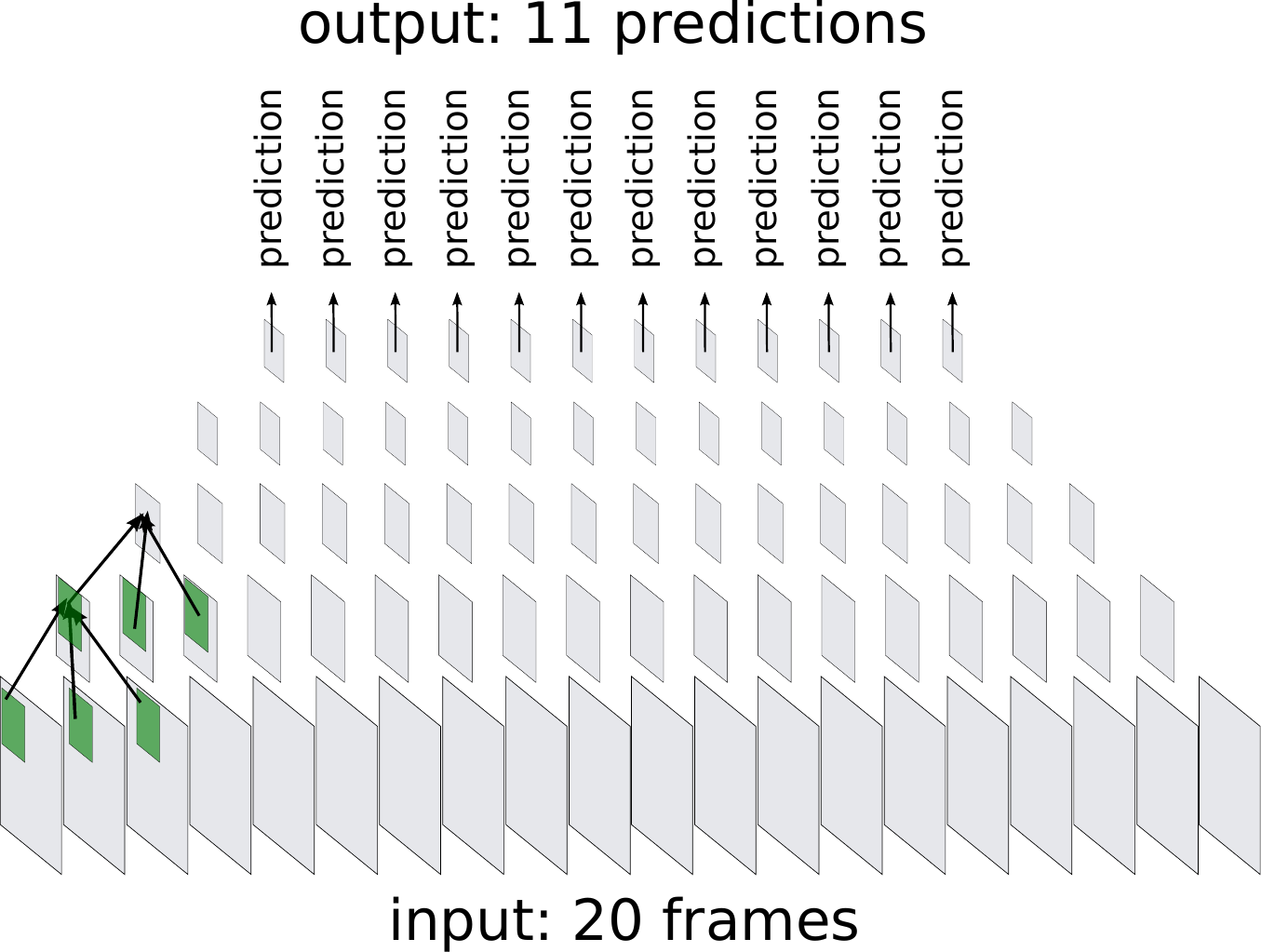} 
  \caption{Our network architecture. Each frame-prediction is based on a context of 10 frames.
  By using a model that is fully convolutional in time, we can increase the input size and thus make~\eg 11 predictions by analyzing 20 frames or 91 predictions by analyzing 100 frames, etc.,
  thus minimizing redundant computation.}
  \label{fig:fully_conv}
\end{figure}

Concurrently to our work, Hassanien~\etal~\cite{hassanien2017large} also have used synthetic data and a CNN for shot detection.
Their architecture is however based on~\cite{tran2015learning} and processes 16 frames at a time, with a stride of 8.
Thus, it processes each frame twice, while our fully convolutional architecture avoids that.
This, and our more compact CNN, allow our model to run at 121x real-time speed on a Nvidia K80 GPU, while~\cite{hassanien2017large} runs at 19x real-time, using faster Nvidia Titan X GPUs.
In addition, because~\cite{hassanien2017large} classifies 16 frames at once, it cannot accurately localize shot transitions and thus requires a post-processing step.
Our fully convolutional architecture, on the other hand, predicts frame-accurate labels directly from pixels.
Figure~\ref{fig:fully_conv} illustrates the advantage of our architecture.

\section{Method}
We pose shot boundary detection as a binary classification problem.
The objective is to correctly predict if a frame is part of the same shot as the previous frame or not.
From this output, it is trivial to obtain the final shot boundaries: We simply assign all frames that are labelled as ``same shot'' to the same shot as the previous frame\footnote{This is in contrast 
to~\cite{hassanien2017large}, which uses an additional SVM and false positive suppression based on color histogram difference.}.

\paragraph{Network architecture.}
To solve the classification problem described above, we propose to use a Convolutional Neural Network.
We use spatio-temporal convolutions to allow the network to analyze changes over time.
The network takes 10 frames as input, runs them through four layers of 3D convolutions, each followed by a ReLU, and finally classifies if the two center frames come from the same shot or
if there is an ongoing transition.
Using 10 frames as input provides context around the frames of interest, something that is important to correctly detect slow transitions such as dissolves.
We use a small input resolution of 64x64 RGB frames for efficiency and since such low resolution are often sufficient for scene understanding~\cite{torralba200880}.
Our network consists of only 48698 trainable parameters. In Table~\ref{tab:arch} we show its architecture in more detail.

\paragraph{Fully convolutional in time.}
Our architecture is inspired by C3D~\cite{tran2015learning}, but is more compact.
More importantly, rather than using fully connected layers, our model consists of 3D convolutions only, thus making the network fully convolutional in time.
The network is given an input of 10 frames, and trained to predict if frame 6 is part of the same shot as frame 5.
Due to its fully convolutional architecture, however, it also accepts larger temporal inputs. E.g. by providing 20 frames, the network would predict labels for frames 6 to 16,
thus making redundant computation unnecessary (also see Figure~\ref{fig:fully_conv}).
This allows to obtain large speedups at inference, as we are showing in our experiments.

\paragraph{Implementation details.}
To train our model we use a cross-entropy loss, which we minimize with vanilla stochastic gradient descent (SGD).
Our model is implemented in TensorFlow~\cite{abadi2016tensorflow}.
At inference, we process the video in snipets of 100 frames, with an overlap of 9 frames.

If a frame is part of a transition such as a dissolve, it is labelled as \textit{not} the same shot as the previous, as it is part of a transition, not a shot.
We found the learning to be more stable when using these kind of labels compared to predicting if a frame is from a new shot or part of a transition.

\begin{table}[t]
\centering
{\renewcommand{\arraystretch}{1.15}
\begin{tabular}{|c|c|c|} 
\hline
\textbf{Layer} & \textbf{Kernel size}  & \textbf{Feature Map} \\
	       &  (w, h, t)  & (w, h, t, channels) \\
\hline
Data & - &  64 x 64 x (10 + $n$) x 3 \\
Conv1 & 5 x 5 x 3 & 30 x 30 x (8 + $n$) x 16 \\
Conv2 & 3 x 3 x 3 & 12 x 12 x (6 + $n$) x 24 \\
Conv3 & 3 x 3 x 3 & 6 x 6 x (4 + $n$) x 32 \\
Conv4 & 6 x 6 x 1 & 1 x 1 x (4 + $n$) x 12 \\
Softmax & 1 x 1 x 4 & 1 x 1 x (1 + $n$) x 2 \\
\hline
\end{tabular}
}
\caption{Our network architecture. All our layers are convolutional and each is followed by a ReLU non-linearity.
By using a fully convolutional architecture, we are able to increase the input size by $n$, thus reusing the shared parts of the convolutional feature map and improving efficiency.}
\label{tab:arch}
\end{table}

\section{Training Data}
\begin{figure}
\includegraphics[width=0.32\linewidth]{images/lOWyng/00049.png} 
 \includegraphics[width=0.32\linewidth]{images/lOWyng/00050.png} 
 \includegraphics[width=0.32\linewidth]{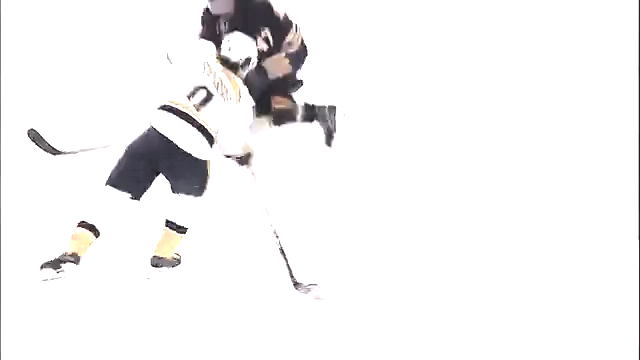} 
  \caption{Artificially generated flash compared a real flash. \textbf{Left:} Input frame. \textbf{Middle:} Frame with artificial flash. \textbf{Right:} The next frame which has a real flash.}
  \label{fig:flash}
\end{figure}
To obtain a dataset large enough to train an end-to-end model we create a new dataset automatically.
The dataset consists of 79 videos with few or no shots transitions and has a total duration of 3.5 hours.
From this data we sample snippets of 10 frames, which will serve as the input to our model. 
To generate training data we combine some of these snippets with a transition.
Thus, we have two types of training examples: (i) snippets consisting of frames from a single shot, i.e. non-transitions and
(ii) transition snippets, which have a transition from one shot to another.

\paragraph{Generated transitions.}
We generate the following transitions:
Hard cuts, Crop cuts (a hard cut to a zoomed in version of the same scene), dissolves, fade-ins, fade-outs and wipes.
Dissolves linearly interpolate between shots, while fades linearly interpolate from or to a frame of a single color.
In wipes a shot is moved out while the next shot is moved in, typically in horizontal direction.
We show the used transitions and their parameters in Table~\ref{tab:transitions}.

\begin{table}[t]
\centering
{\renewcommand{\arraystretch}{1.15}
\begin{tabular}{|c|c|c|} 
\hline
\textbf{Transition} & \textbf{Duration} & \textbf{Other} \\
\hline
Cuts & 1 frame & - \\
\hline
Crop cuts & 1 frame & Crop to 50-70\% of full size \\
\hline
Dissolves & 3 to 14 frames & -\\
\hline
Fade in/out & 3 to 14 frames & fade to/from black or white \\
\hline
Wipes & 6 to 9 frames & horizontal direction \\
\hline
\end{tabular}
}
\caption{The generated transitions and their parameters.}
\label{tab:transitions}
\end{table}

\paragraph{Flashes.}
Videos often contain flashes (\cf Figure~\ref{fig:hard_cases_flash}), which result in heavy frame-to-frame changes.
This has typically posed problems to SBD methods, which were required to remove these false positives in a post-processing step~\cite{smeaton2010video}.
Instead, we choose a different approach: We add artificial flashes to the non-transition snippets to make the network invariant to these kind of changes.
For this purpose we transform random frames by converting them to the LUV color space and increasing the intensity by 70\%.
Figure~\ref{fig:flash} shows an example result of this procedure, compared to real flash.

\section{Experiments}

\begin{table}[t]
\centering
{\renewcommand{\arraystretch}{1.15}
\begin{tabular}{|l|l|l|} 
\hline
& \textbf{F1 score} & \textbf{Speed} \\
\hline
Apostolidis et al.~\cite{apostolidis2014fast} & 0.84         & 3.3x (GPU)      \\
Baraldi et al.~\cite{baraldi2015shot}   & 0.84         & 7.7x (CPU)      \\
Song et al.~\cite{song2016click}        & 0.68        & 33.2x (CPU)        \\ 
\hline 
Ours        & \textbf{0.88}         & \textbf{121x (GPU)}        \\ 
w/o fully conv. inference       &  & 13.9x (GPU)        \\
\hline
\end{tabular}
}
\caption{Performance and speed comparison on the RAI dataset~\cite{baraldi2015shot}.
As can be seen our method significant outperforms previous works, while being significantly faster.
We note, however, that~\cite{baraldi2015shot} uses a single-threaded CPU implementation, while we run our method on a GPU.}
\label{fig:rai_performance}
\end{table}
We evaluate our method on the publicly available RAI dataset~\cite{baraldi2015shot}.
Thereby we follow~\cite{baraldi2015shot} and report F1 scores. It is computed as the harmonic 
mean between transition recall and precision. \ie it measures how many shot changes are correctly detected or missed and how many false positives a method has.
Table~\ref{fig:rai_performance} summarizes the results.
We now discuss these results in detail.

\begin{figure}
  \begin{subfigure}[t]{1\linewidth}
\includegraphics[width=0.32\linewidth]{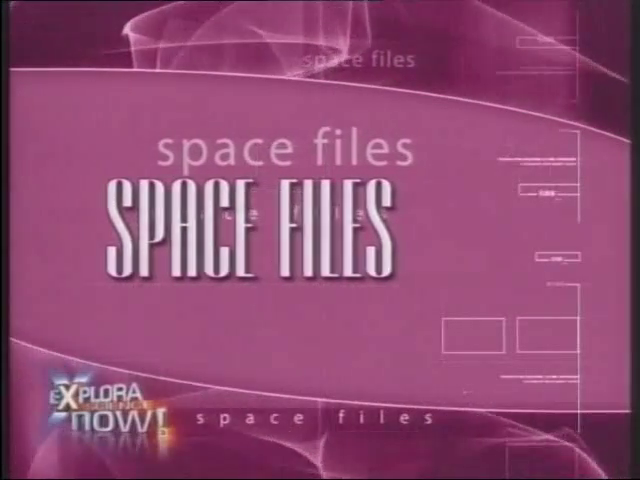} 
 \includegraphics[width=0.32\linewidth]{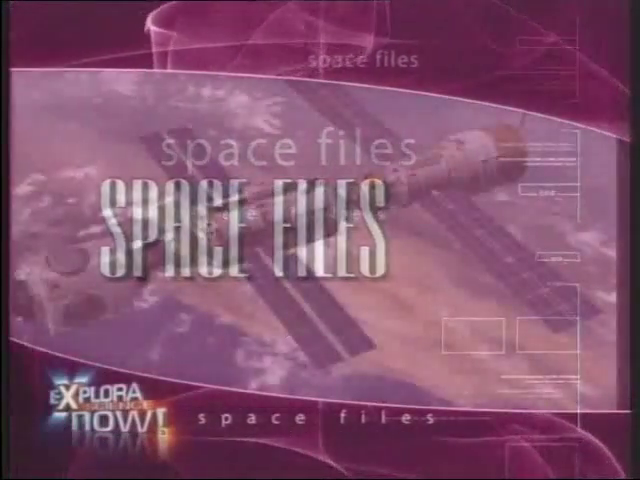} 
 \includegraphics[width=0.32\linewidth]{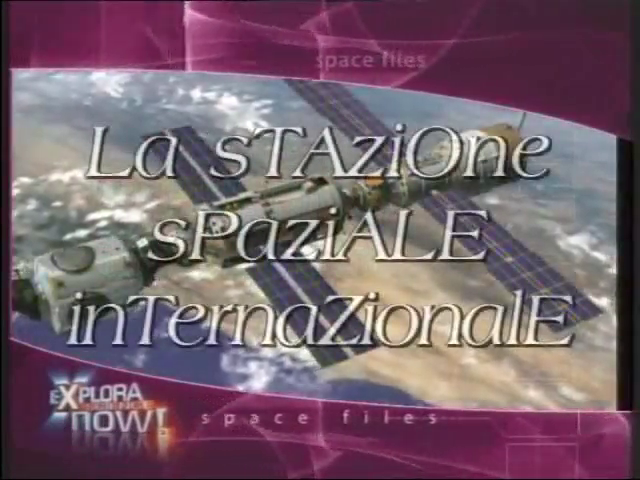} 
    \caption{Missed partial dissolve (annotated as a transition).}
    \label{fig:amb_1}
    \end{subfigure}
  \begin{subfigure}[t]{1\linewidth}
  \centering
\includegraphics[width=0.32\linewidth]{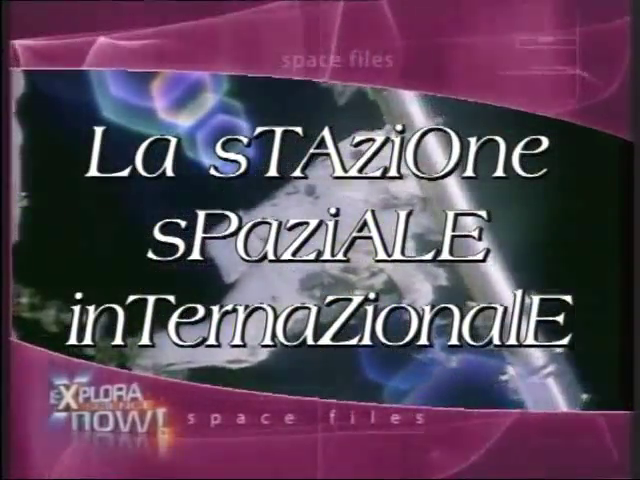} 
 \includegraphics[width=0.32\linewidth]{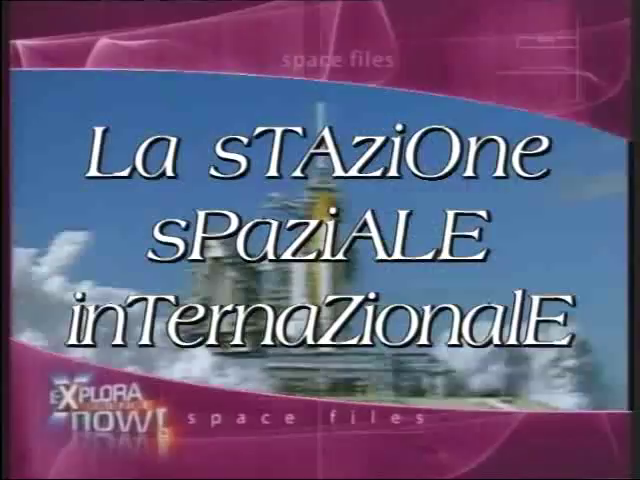} 
    \caption{Falsely detected partial hard cut (annotated as no transition).}
    \label{fig:amb_2}
    \end{subfigure}    
  \begin{subfigure}[t]{1\linewidth}
\includegraphics[width=0.32\linewidth]{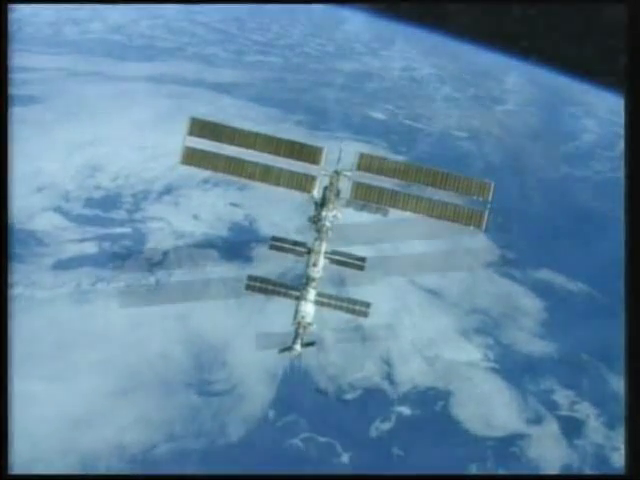} 
 \includegraphics[width=0.32\linewidth]{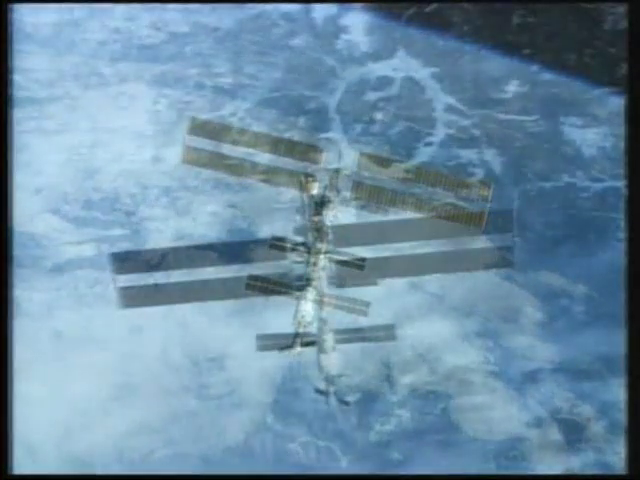} 
 \includegraphics[width=0.32\linewidth]{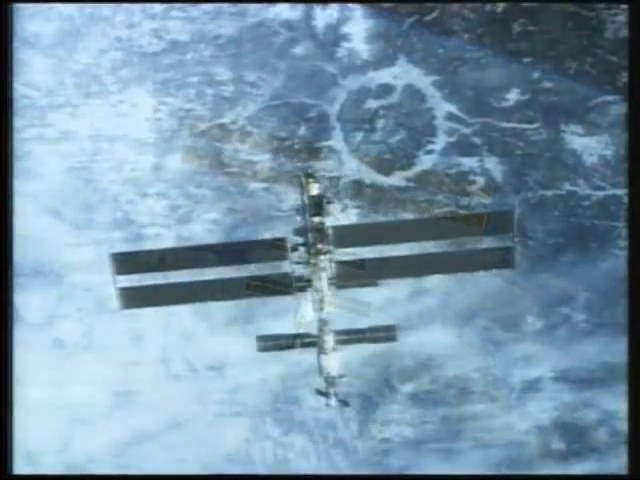} 
    \caption{Missed slow dissolve (first and last frame are 1 second apart)}
    
    \end{subfigure}
  \begin{subfigure}[t]{1\linewidth}
  \centering
\includegraphics[width=0.32\linewidth]{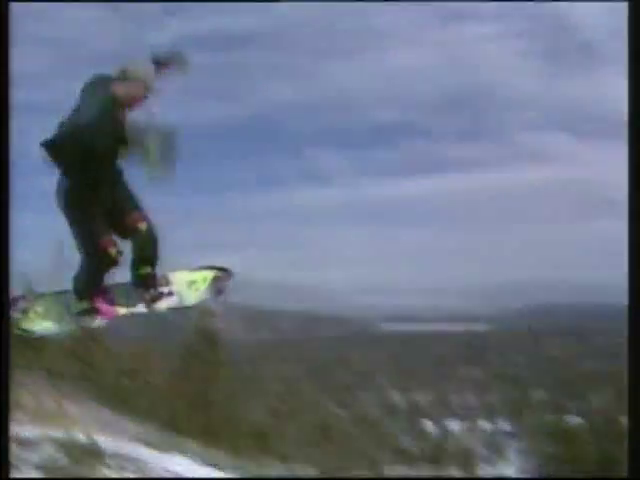} 
 \includegraphics[width=0.32\linewidth]{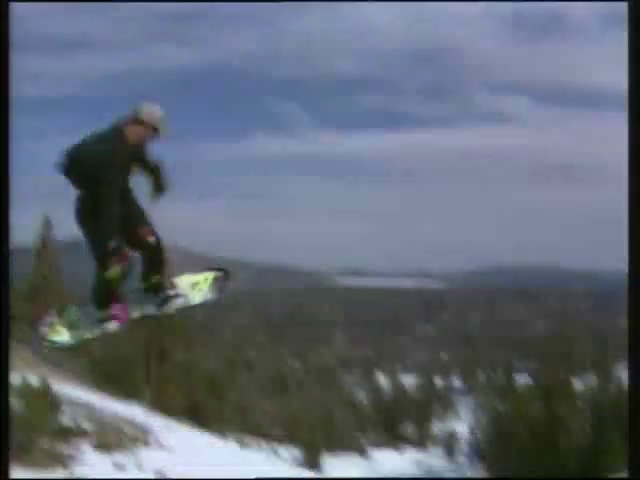} 
    \caption{False positive due to fast motion with motion blur)}
    \end{subfigure}    
  \begin{subfigure}[t]{1\linewidth}
  \centering
\includegraphics[width=0.32\linewidth]{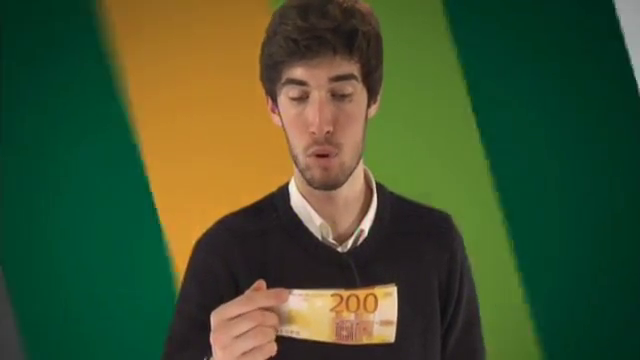} 
 \includegraphics[width=0.32\linewidth]{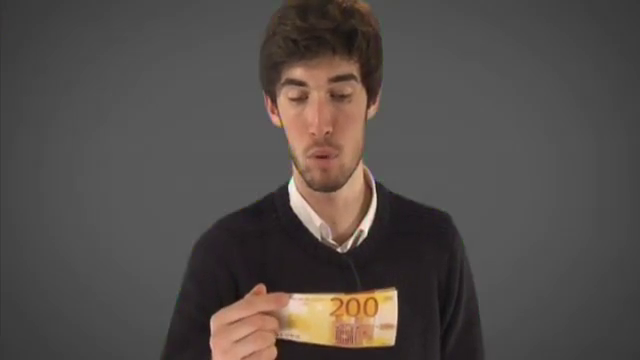} 
    \caption{Falsely detected partial hard cut (annotated as no transition).}
    \label{fig:partial_cut}
    \end{subfigure}    
    
  \caption{Error cases on the RAI dataset~\cite{baraldi2015shot}.
  Our method makes most errors on partial scene changes, where labelling a shot change depends on what is fore- and background, and has problems with fast motion.
  We also note that some cases are ambiguous, \eg transitions in (a) and (b).}
  \label{fig:errors}
\end{figure}

\paragraph{Performance.}
As can be seen from Table~\ref{fig:rai_performance}, our method outperforms the previous state of the art methods on this dataset.
Our method improves the mean F1 score from 84\% to 88\%, thus reducing the errors by 25\%.
It obtains an accuracy of more than 90\% in recall and precision on most videos (\cf Table~\ref{tab:detail_rai}).
Lower precision stems from custom transitions such as partial cuts as shown in Figure~\ref{fig:partial_cut}).
Indeed, more than 90\% of the false positives on videos 25011 and 25012 are due to these effects.
These are however special cases and extremely hard to detect. First, if such partial transitions are considered a shot change typically
depends on whether the foreground or background changes. Second, such transitions were not included into training.
On common transitions such as hard cuts, fades and dissolves our method is extremely accurate.

\begin{table}[t]
\centering
\label{my-label}
{\renewcommand{\arraystretch}{1.15}
\begin{tabular}{|c|c|c|} 
\hline
\textbf{Video} & \textbf{Precision}  & \textbf{Recall} \\
\hline
21867 & 0.89 & 0.84 \\
23553 & 0.95 & 0.99 \\
23557 & 0.91 & 0.97 \\
23558 & 0.92 & 0.99 \\
25008 & 0.94 & 0.94 \\
25009 & 0.97 & 0.96 \\
25010 & 0.93 & 0.94 \\
\hline
21829 & 0.81 & 0.64 \\
25011 & 0.62 & 0.90 \\
25012 & 0.66 & 0.89 \\
\hline
\end{tabular}
}
\caption{Per video results on the RAI dataset. We obtain Recall and Precision $>0.9$ on most videos.
Videos with lower precision or recall have custom transitions or animations. See Figure~\ref{fig:errors} for examples.}
\label{tab:detail_rai}
\end{table}

\paragraph{Speed comparison.}
For measuring speed, we use a machine with an single Nvidia K80 GPU and 32 Intel Xeron CPUs with 2.30GHz.
We measured the speed of~\cite{song2016click} and our method on this machine, using the video 25012 of the RAI dataset.
For~\cite{apostolidis2014fast,baraldi2015shot} we used the timings provided in the paper instead.
Thus, these numbers don't allow for an exact comparison, but rather give a coarse indication of the relative speed of the different methods.

From Table~\ref{fig:rai_performance} we can see that our method is much faster than previous methods.
We also show that making the architecture fully convolutional is crucial for obtaining fast inference speed.
To the best of our knowledge, our model it is the fastest shot detection to date.
Indeed, as our network uses a small input resolution of 64x64, the current bottleneck is not the network itself, but rather bi-linear resizing in ffmpeg.
When we resize the video to 64x64 prior to running our model, it obtains 235x real-time or $\approx$5895 FPS.

\section{Conclusion}
In this paper we have introduced a novel method for shot boundary detection.
Our CNN architecture is fully convolutional in time, thus allowing for shot detection at more than 120x real-time.
We have compared our model against the state-of-the art on the RAI dataset~\cite{baraldi2015shot} and have shown that it outperforms previous works, while requiring a fraction of time.
We also note that our model was not using any real-world shot transitions for training.
Currently, our model makes three main errors, which we visualize in Figure~\ref{fig:errors}: (i) missing long dissolves, which it was not trained with, (ii) partial scene changes and 
(iii) fast scenes with motion blur.
In the future, we would like to include real data into training, such that the network can learn about rare and unusual shot transitions and become 
more robust to rapid scene progression.
Furthermore we will evaluate our method on the TRECVid that, in order to be able to compare to~\cite{hassanien2017large}.

\section{Acknowledgment}
I would like to thank the gifs.com team for useful discussions and support, as well as Eirikur Agustsson, Anna Volokitin, Jordi Pont-Tuset, Santiago Manen.

{\small
\bibliographystyle{ieee}
\bibliography{references}

\begin{thebibliography}{10}\itemsep=-1pt

\bibitem{abadi2016tensorflow}
M.~Abadi, P.~Barham, J.~Chen, Z.~Chen, A.~Davis, J.~Dean, M.~Devin,
  S.~Ghemawat, G.~Irving, M.~Isard, et~al.
\newblock Tensorflow: A system for large-scale machine learning.
\newblock In {\em OSDI}, 2016.

\bibitem{akutsu1992video}
A.~Akutsu, Y.~Tonomura, H.~Hashimoto, and Y.~Ohba.
\newblock Video indexing using motion vectors.
\newblock In {\em Applications in Optical Science and Engineering}, 1992.

\bibitem{apostolidis2014fast}
E.~Apostolidis and V.~Mezaris.
\newblock Fast shot segmentation combining global and local visual descriptors.
\newblock In {\em ICASSP}, 2014.

\bibitem{baraldi2015shot}
L.~Baraldi, C.~Grana, and R.~Cucchiara.
\newblock Shot and scene detection via hierarchical clustering for re-using
  broadcast video.
\newblock In {\em CAIP}, 2015.

\bibitem{bay2008speeded}
H.~Bay, A.~Ess, T.~Tuytelaars, and L.~Van~Gool.
\newblock {Speeded-up robust features (SURF)}.
\newblock {\em CVIU}.

\bibitem{cernekova2006information}
Z.~Cernekova, I.~Pitas, and C.~Nikou.
\newblock Information theory-based shot cut/fade detection and video
  summarization.
\newblock {\em Circuits and systems for video technology}, 2006.

\bibitem{gygli2015video}
M.~Gygli, H.~Grabner, and L.~Van~Gool.
\newblock Video summarization by learning submodular mixtures of objectives.
\newblock In {\em CVPR}, 2015.

\bibitem{hassanien2017large}
A.~Hassanien, M.~Elgharib, A.~Selim, M.~Hefeeda, and W.~Matusik.
\newblock Large-scale, fast and accurate shot boundary detection through
  spatio-temporal convolutional neural networks.
\newblock {\em arXiv preprint arXiv:1705.03281}, 2017.

\bibitem{hitchcock2001rope}
A.~Hitchcock, J.~Stewart, J.~Dall, F.~Granger, A.~Laurents, and P.~Hamilton.
\newblock {\em Rope}.
\newblock Universal Studios, 2001.

\bibitem{long2015fully}
J.~Long, E.~Shelhamer, and T.~Darrell.
\newblock Fully convolutional networks for semantic segmentation.
\newblock In {\em CVPR}, 2015.

\bibitem{smeaton2010video}
A.~F. Smeaton, P.~Over, and A.~R. Doherty.
\newblock Video shot boundary detection: Seven years of trecvid activity.
\newblock {\em CVIU}, 2010.

\bibitem{song2016click}
Y.~Song, M.~Redi, J.~Vallmitjana, and A.~Jaimes.
\newblock To click or not to click: Automatic selection of beautiful thumbnails
  from videos.
\newblock In {\em CIKM}. ACM, 2016.

\bibitem{torralba200880}
A.~Torralba, R.~Fergus, and W.~T. Freeman.
\newblock 80 million tiny images: A large data set for nonparametric object and
  scene recognition.
\newblock {\em PAMI}, 2008.

\bibitem{tran2015learning}
D.~Tran, L.~Bourdev, R.~Fergus, L.~Torresani, and M.~Paluri.
\newblock Learning spatiotemporal features with 3d convolutional networks.
\newblock In {\em ICCV}, 2015.

\bibitem{wang2017untrimmednets}
L.~Wang, Y.~Xiong, D.~Lin, and L.~Van~Gool.
\newblock Untrimmednets for weakly supervised action recognition and detection.
\newblock {\em CVPR}, 2017.

\end{thebibliography}
}
\end{document}